\documentclass[letterpaper, 10 pt, conference]{ieeeconf}   
\usepackage{cite}
\usepackage{amsmath}
\usepackage{amsfonts}
\usepackage{steinmetz}
\usepackage{graphicx}
\usepackage{subfigure}
\usepackage{adjustbox}
\usepackage{multirow}
\usepackage{colortbl}
\usepackage{caption}
\usepackage{hyperref}
\usepackage{fancyhdr}
\newcommand\kevin[1]{\textcolor{black}{#1}}
\newcommand\llz[1]{\textcolor{black}{#1}}

\IEEEoverridecommandlockouts                              

\title{\LARGE \bf
NDT-Map-Code: A 3D global descriptor for real-time loop closure detection in lidar SLAM
}

\author{Lizhou Liao$^{1,\dagger}$, Wenlei Yan$^{1,\dagger}$, Li Sun$^{2,3*}$, Xinhui Bai$^{2}$, Zhenxing You$^{2}$, Hongyuan Yuan$^{2}$, Chunyun Fu$^{1,4}$
\thanks{This work was supported by the Chongqing Technology Innovation and Application Development Project under Grant CSTB2022TIAD-DEX0013.}
\thanks{$\dagger$These two authors contribute equally to this work.}
\thanks{$^{1}$The College of Mechanical and Vehicle Engineering, Chongqing University, China. E-mail: 
        {\tt\small liaolizhou@cqu.edu.cn, yanwenlei@stu.cqu.edu.cn, fuchunyun@cqu.edu.cn}.}%
\thanks{$^{2}$The autonomous driving division, NIO.  E-mail: 
        {\tt\small kevin.sun, xinhui.bai, zhenxing.you, hongyuan.yuan@nio.com}.}%
\thanks{$^{3}$Department of Computer Science, The University of Sheffield, UK}
\thanks{$^{4}$The State Key Lab of Mechanical Transmissions, Chongqing University, China.}
\thanks{$^{*}$The corresponding author.}
}

\begin{document}

\maketitle

\setlength{\voffset}{-10mm} 
\setlength{\headsep}{5mm}
\thispagestyle{fancy}

\pagestyle{empty}

\begin{abstract}

\kevin{Loop-closure detection, also known as place recognition, aiming to identify previously visited locations, is an essential component of a SLAM system. Existing research on lidar-based loop closure heavily relies on dense point cloud and 360 FOV lidars. This paper proposes an out-of-the-box NDT (Normal Distribution Transform) based global descriptor, NDT-Map-Code, designed for both on-road driving and underground valet parking scenarios. NDT-Map-Code can be directly extracted from the NDT map without the need for a dense point cloud, resulting in excellent scalability and low maintenance cost. The NDT representation is leveraged to identify representative patterns, which are further encoded according to their spatial location (bearing, range, and height). Experimental results on the NIO underground parking lot dataset and the KITTI dataset demonstrate that our method achieves significantly better performance compared to the state-of-the-art.}

\end{abstract}

\section{INTRODUCTION}
\kevin{Loop-closure detection is the key technique for eliminating the long-term drift in large-scale mapping when the GPS is not available (e.g. automated valet parking). Moreover, SLAM methods for large-scale applications require good adaptations of using lightweight maps since the trend of mapping will be based on onboard computation and crowd-sourced data. Existing lidar-based loop closure detection methods\cite{kim2018scan, wang2020lidar, li2021ssc, wang2020intensity} usually convert a frame of point cloud into a two-dimensional global descriptor through polar-coordinate ROI partitioning, and, the rotation invariance can be achieved by column-wise shifting. The main-stream methods have two limitations: firstly, the existing methods require detailed point-level geometry information hence dense point cloud map, and 360 FOV lidar scans are necessarily used for feature representation. This will boost the requirements for onboard storage and vehicle-cloud data transmission; secondly, in contrast to on-road driving scenarios, significant adaptation is required to deal with repetitive patterns, dynamic objects, and occlusion in structures for underground parking-lots localization and mapping.}

\kevin{This paper proposes a global descriptor, NDT-Map-Code (NDT-MC), that is highly complementary with scalable NDT point clouds built through a crowd-sourced mapping way.  The proposed method is devised to discover and describe the structural landmarks in consideration of NDT cells' geometrical shape, entropy, and spatial context.  Our intuition is to describe the place by `what' landmarks at `where'. To describe `what', we classify the geometric shapes of NDT cells. The entropy of the chosen NDT cells is considered to shortlist effective geometric patterns. To describe `where', we propose a polar-range-height-coordinate-based ROI partitioning. Instead of focusing on structures of the largest height, we divide the entire surrounding scene structure into multiple layers according to height. Afterward, both NDT's shape types and their heights are employed to formulate a multi-layer global descriptor. For front-view lidars, we employ lidar odometry to construct and maintain sub-maps. This approach can improve the limited co-visibility of the front-view-only LiDARs.}

The main contributions of our approach are:
\begin{itemize}
   \item
    \llz{A novel global descriptor, NDTMC, for both underground parking scenarios and on-road driving scenes is proposed.  This descriptor is complementary to scalable NDT representation, which utilizes geometric and  entropy patterns in NDT and represents of scene features through multi-layer shape and context encoding;}
   \item
   \kevin{Extensive experiments conducted on NIO underground parking-lot dataset (i.e. a collection of eight sequences in four parking lots), coupled with six sequences in the KITTI dataset, underpin the superiority of our method over state-of-the-art approaches;}
   \item 
   \kevin{We made the proposed method, together with an integrated real-time full-SLAM system, which is named NDTMC-LIO-SAM, publicly accessible to the community, contributing to its potential benefits, and is subject to appropriate license agreements \url{https://github.com/SlamCabbage/NDTMC}.}
\end{itemize}

\section{RELATED WORK}

\kevin{Existing research on loop closure includes lidar-based methods and vision-based methods. Lidar-based methods have received increasing attention due to their robustness to lightness and illumination variance. A number of lidar-based loop closure methods have been proposed in recent years.}
\kevin{A stream of global descriptors can be constructed as a global statics of 3D local descriptors.
Several keypoint detection methods have been used, such as 3D Sift\cite{3DSIFT}, Link3D\cite{link3d}, 3D-SURF\cite{3dsurf}, BoW3D\cite{bow3d}, SHOT\cite{SHOT}, and Imaging-Lidar\cite{ImagingLidar}. After that, the methods of voting\cite{vote} and Bag-of-Words\cite{bow} are used to combine these descriptors for loop closure.}

\begin{figure}[t]
\centering  
\includegraphics[width=0.98\linewidth]{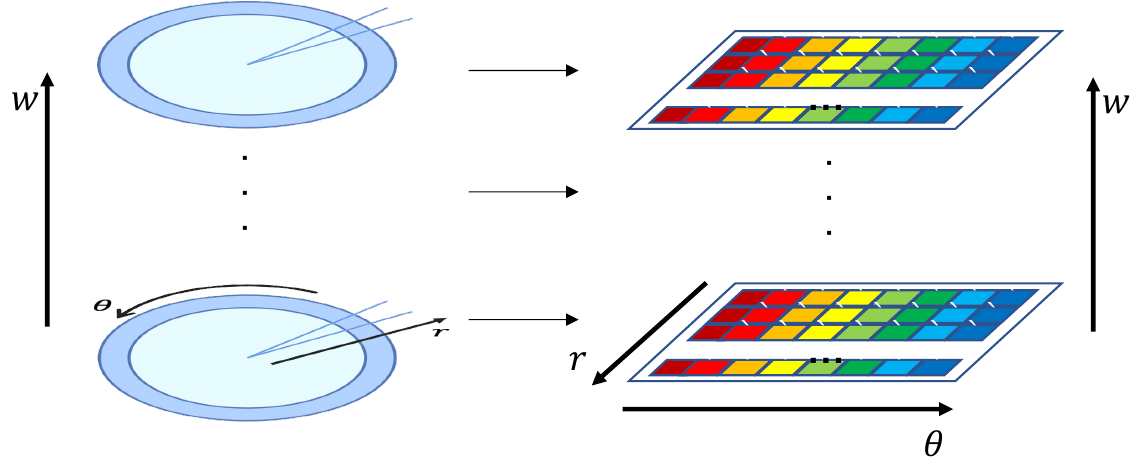} 
\caption{\kevin{Polar-range-height coordinates based ROI partitioning. We first divide the 3D space according to ring, sector, and height, and the corresponding polar-range-height coordinate axes, i.e. $r$, $\theta$, and $w$, can be obtained. Afterward, the ring sector corresponding to different heights ($w$) is transformed into a $Cartesian$ coordinate system with $\theta$ as the abscissa and $r$ as the ordinate.}}
\label{TE}
\vspace{-0.2in} 
\end{figure}
\kevin{Compared with local descriptor methods, global-based descriptors have better reproducibility in changing scenarios \cite{locus}. More recently, M2DP\cite{M2DP} generates a vector descriptor by projecting a point cloud onto a plane at multiple angles. Since this method's Principal Component Analysis (PCA) of its point cloud does not guarantee that the two point clouds can be robustly aligned, the M2DP descriptor lacks rotation invariance. Methods \cite{kim2018scan,li2021ssc,wang2020intensity} perform polar-coordinate ROI partitioning on the bird-eye viewpoint cloud to generate a two-dimensional descriptor, which can achieve rotation invariance between two descriptors through matrix shift. In \cite{kim2018scan}, each bin of the feature matrix retains the maximum height to represent all points in the bin; \cite{li2021ssc} and \cite{wang2020intensity} include the semantic labels and maximum intensity values respectively, apart from height values. Lidar Iris\cite{wang2020lidar} divides the vertical FOV into multiple layers according to the pitch angle of the point, and each layer uses the same polar-range coordinates partition as \cite{kim2018scan}. The value of a bin is assigned 0 or 1 depending on whether it is occupied or not. Finally, binary-encoded of all bins with different layers and the same polar-range partition to get the final Lidar Iris descriptor. The core idea of the Normal Distribution Descriptor(NDD)\cite{NDD} method is to map point cloud data to the Range-Polar coordinate system and calculate the mean and covariance of points in the bins in each coordinate system. Through these means and covariances, NDD calculates two important descriptors, namely probability density score and entropy, which have different properties. Finally, these two descriptors are fused into a complete NDD descriptor.}

\kevin{Recently, deep learning methods \cite{locus, ma2022overlaptransformer, dewan2018learning, yin2017efficient, dube2020segmap, uy2018pointnetvlad, arandjelovic2016netvlad, NDT-Transformer} have been used to learn feature descriptors, and these methods have shown significant performance improvements compared to traditional methods. However, learning-based methods show limited generalizability for novel scenes, and high computation resources are required for deployment.}

\kevin{From the literature, existing methods have three limitations: 1) firstly, most of the above methods are devised based on 360-degree lidars, and it is likely to fail with a front-view lidar; 2) secondly, due to limitations of public datasets, existing descriptors are primarily designed for outdoor on-road driving rather than underground parking scenes;} 3) finally, existing descriptors require dense point cloud maps for feature extraction, which does not suit lightweight maps obtained by crowd-sourced mapping.

\begin{table}[b]
\centering
\caption{Comparison of storage space size of point cloud types}
\label{ndtsize}
\resizebox{0.5\textwidth}{!}{
\begin{tabular}{l|l|l}
~             & Raw
  Point Cloud (MB)                & NDT map with 2 m resolution (MB) \\ 
\hline
Rongke        & 2252.8   & 9.1 \\
Yinwang & 889.7    & 1.7  \\
Yinzuo        & 1223.7   & 4.8 \\
Lixiangguoji  & 1331.2   & 4.9           
\end{tabular}}      
\end{table}

\begin{figure*}[t]
 \centering  
 \includegraphics[width=0.98\textwidth]{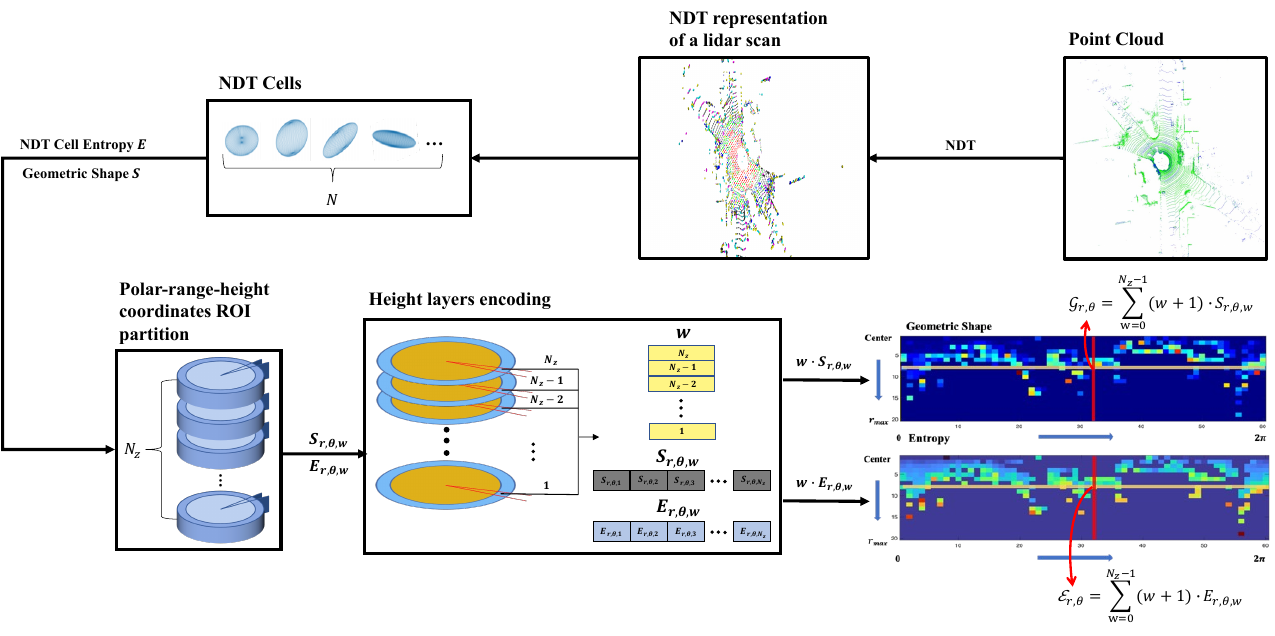} 
 \caption{The overview of NDT-MC. The construction of NDT-MC involves the following steps: 1. Converting point clouds into NDT representation. 2. Calculating geometric values and entropy for each NDT cell. 3. Computing the coordinates of each NDT cell in the Polar-Range-Height coordinate system based on its mean value. 4. Constructing the proposed descriptor using a strengthened height-layer encoding with geometric and entropy components.}
 \label{pipeline}
\end{figure*}

\section{METHODOLOGY}

\begin{figure}
 \centering
 \begin{minipage}{1\linewidth} 
  \subfigure[plane ($0 < g \le 0.3$) ]{
   \label{g_1}
   \includegraphics[width=0.49\linewidth,height=1.2in]{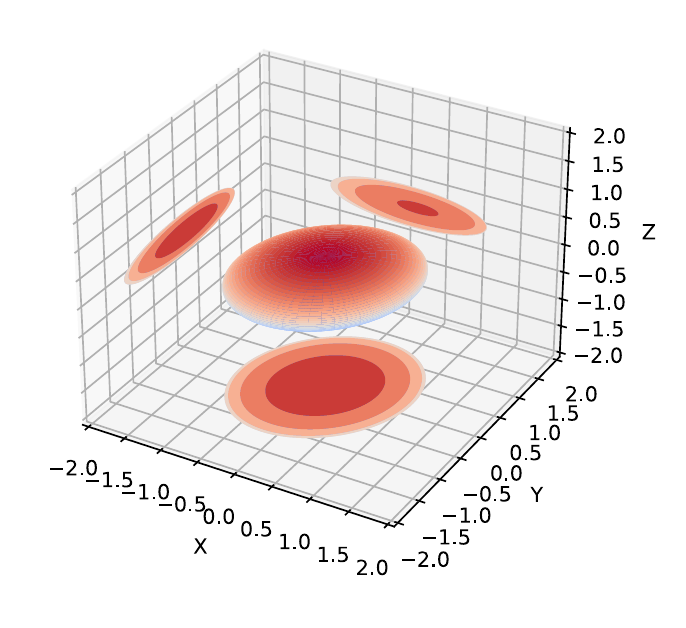} 
  }\noindent
  \subfigure[ellipsoid ($0.3 < g \le 0.7$)]{
   \label{g_2}
   \includegraphics[width=0.49\linewidth,height=1.2in]{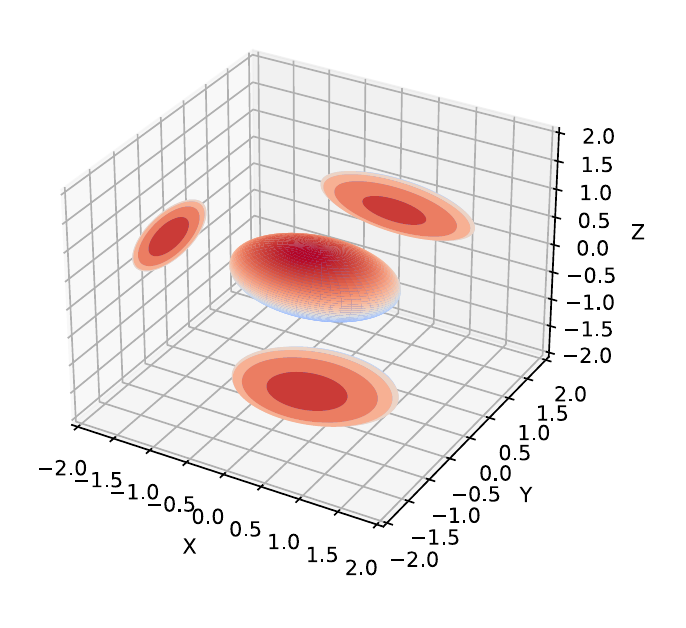}
  }
 \end{minipage}
 \vskip -0.3cm 
 \begin{minipage}{1\linewidth }
  \subfigure[sphere ($0.7 < g \le 2$)]{
   \label{g_3}
   \includegraphics[width=0.49\linewidth,height=1.2in]{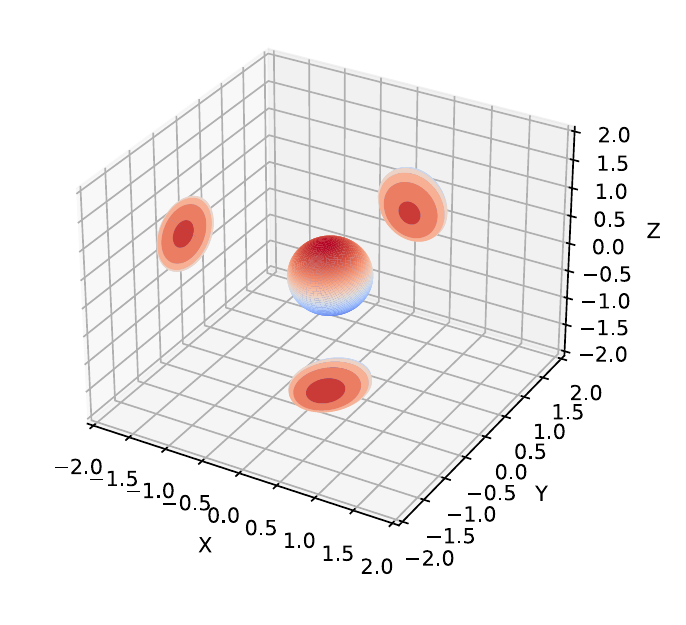}
  }\noindent
  \subfigure[line ($2 < g \le 8$)]{
   \label{g_4}
   \includegraphics[width=0.49\linewidth,height=1.2in]{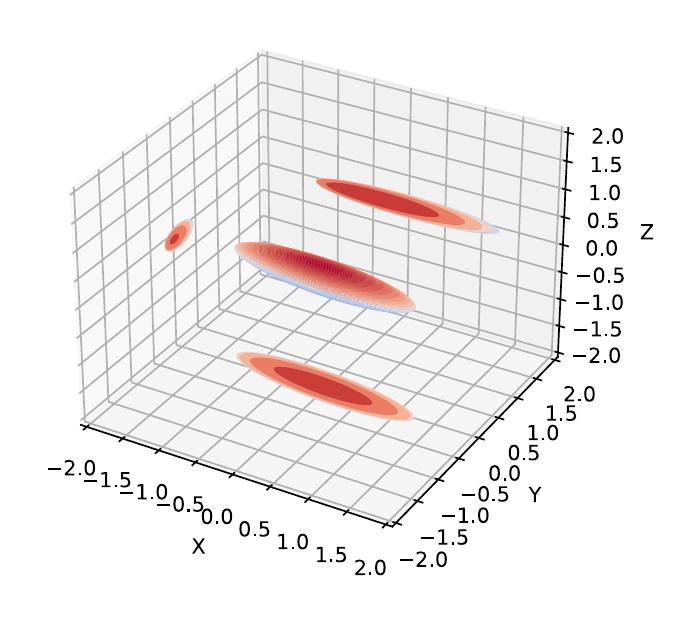}
  }
 \end{minipage}
 \caption{\kevin{The geometrical shape of NDT cells corresponding to different $g$ values. Four subfigures illustrate representative shapes for corresponding $g$ values, namely plane, ellipsoid, sphere, and line. In each subfigure, a 3D NDT cell is shown along with a 2D projection in the X, Y, and Z directions.}}
 \label{g_figure}
\end{figure}

\begin{figure}[t] 
\centering 
\includegraphics[height=5cm]{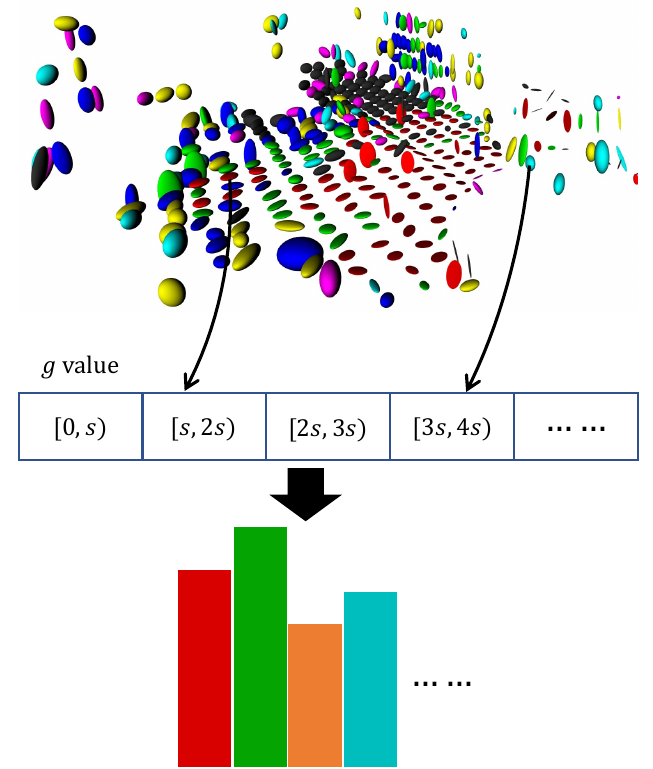} 
\caption{Construction of geometric key. $s$ is a segmented parameter for the $g$ value.} 
\label{GK} 
 \vspace{-0.2in}
\end{figure}

\subsection{NDT represent point cloud}
NDT is used to represent the point cloud due to its scalability in mapping large-scale scenes, as well as its advantages in noise and outlier removal and reducing the number of points to improve processing speed. 
As indicated in Table \ref{ndtsize}, the storage space needed for a 2m resolution NDT map accounts for merely 0.36\% of the storage space required by the raw point cloud. To obtain a point cloud in NDT representation, we divide the point cloud into uniformly distributed 3D grids and estimate a multivariate Gaussian for each cell. A one-pass algorithm is used to calculate the mean and covariance:
\begin{equation}
\small
\begin{split}
\label{mean}
 (\bar{x}_i,\bar{y}_i,\bar{z}_i) = (\bar{x}_{i-1},\bar{y}_{i-1},\bar{z}_{i-1})+ \\  \frac{1}{i}(x_i-\bar{x}_{i-1},y_i-\bar{y}_{i-1},z_i-\bar{z}_{i-1}) \ \\ 
\end{split}
\end{equation}
\begin{equation}
\small
\begin{split}
\label{C_i}
C_i = C_{i-1} + \frac{i-1}{i}(x_i-\bar{x}_{i-1},y_i-\bar{y}_{i-1},z_i-\bar{z}_{i-1})^T \\ 
\cdot (x_i-\bar{x}_{i-1},y_i-\bar{y}_{i-1},z_i-\bar{z}_{i-1}) \ \\
\end{split}    
\end{equation}
\begin{equation}
\small
\begin{split}
\label{Cov_n}
{Cov}_n = \frac{C_n}{n}
\end{split}
\end{equation}
\kevin{where $\left({\bar{x}}_i,{\bar{y}}_i,{\bar{z}}_i\right)$ refers to the mean value of the previous $i$ points in the same Cell}, and $n$ is the total number of points, $C$ is a matrix of $3\times3$, and ${Cov}_n$ represents the covariance matrix of $n$ points.

\subsection{Polar-range-height coordinates ROI partition}

\kevin{Similar to Scan-Context-like methods, polar-range coordinates ROI partition is used to divide 3D space into rings and sectors:}
\begin{equation}
\small
\label{rlayer}
    r_k=max\left\{r_k\in\mathbb{Z}\ |\ r_k\le\frac{\sqrt{x_k^2+y_k^2}}{L_r},0\le\sqrt{x_k^2+y_k^2}\le R\right\}
\end{equation}
\begin{equation}
\small
    \Theta\left(\frac{y_k}{x_k}\right)=\left\{
    \begin{aligned}
    arctan\frac{y_k}{x_k},{0<x}_k,\ 0\le y_k \\
    arctan\frac{y_k}{x_k}+\pi,x_k < 0 \\
    arctan\frac{y_k}{x_k}+2\pi,0<x_k,y_k<0
    \end{aligned}
    \right.
\end{equation}
\begin{equation}
\small
\label{thetalayer}
    \theta_k=max\left\{\theta_k\in\mathbb{Z}\ |\ \theta_k\le\frac{\Theta\left(\frac{y_k}{x_k}\right)}{L_\theta},0\le\sqrt{x_k^2+y_k^2}\le R\right\}
\end{equation}
\kevin{where $L_r$ and $L_\theta$ represent the radial resolution and angular resolution of polar coordinates respectively and $R$ represents the maximum radial distance allowed. $\theta_k$ and $r_k$ represent the polar and range coordinates indices of $p_k$, whose maximum values are $N_\theta$ and $N_r$ respectively.}
 
\kevin{Existing methods focus on describing the buildings for outdoor localization. Specifically, each bin of \cite{kim2018scan} retains the maximal height value to represent the point cloud in the entire ROI bin; \cite{li2021ssc} and \cite{wang2020intensity} include the most semantic labels and the maximum intensity value, respectively. }

\kevin{In underground parking lots, the presence of dynamic objects, such as parked cars, can significantly impact scan-context-like approaches. Moreover, due to the equivalent height of ceiling,  maximal-height-based descriptors show limited effectiveness in discriminating between different locations. }


\kevin{To address these challenges, as illustrated in Fig. \ref{TE}, we propose an alternative approach to dividing the scene $p_k(x_k, y_k, z_k)$ into multiple layers based on height values with respect to the vehicle's base link. Unlike the method proposed in \cite{wang2020lidar}, we do not use pitch angles for vertical partition. Instead, we segment the scene into layers according to the height values.}

\begin{equation}
\label{zlayer}
    w_k=max\left\{{w_k} \in \mathbb{Z} | {w_k} \leq \frac{z_k}{L_z}, 0 \leq {z_k} \leq Z \right\}
\end{equation}
\kevin{Here, $\mathbb{Z}$ means natural number, $L_z$ represents the height of each layer, $Z$ refers to the maximal $z$ value of truncated points $[0, Z]$, and $w_k$ represents the layer index to which $p_k$ belongs, whose maximum value is $N_w$.}

\subsection{Classification of NDT shapes and calculation of entropy}
\begin{figure*}[t]
    \centering
    \subfigure[Rongke]{\includegraphics[width=0.24\textwidth]{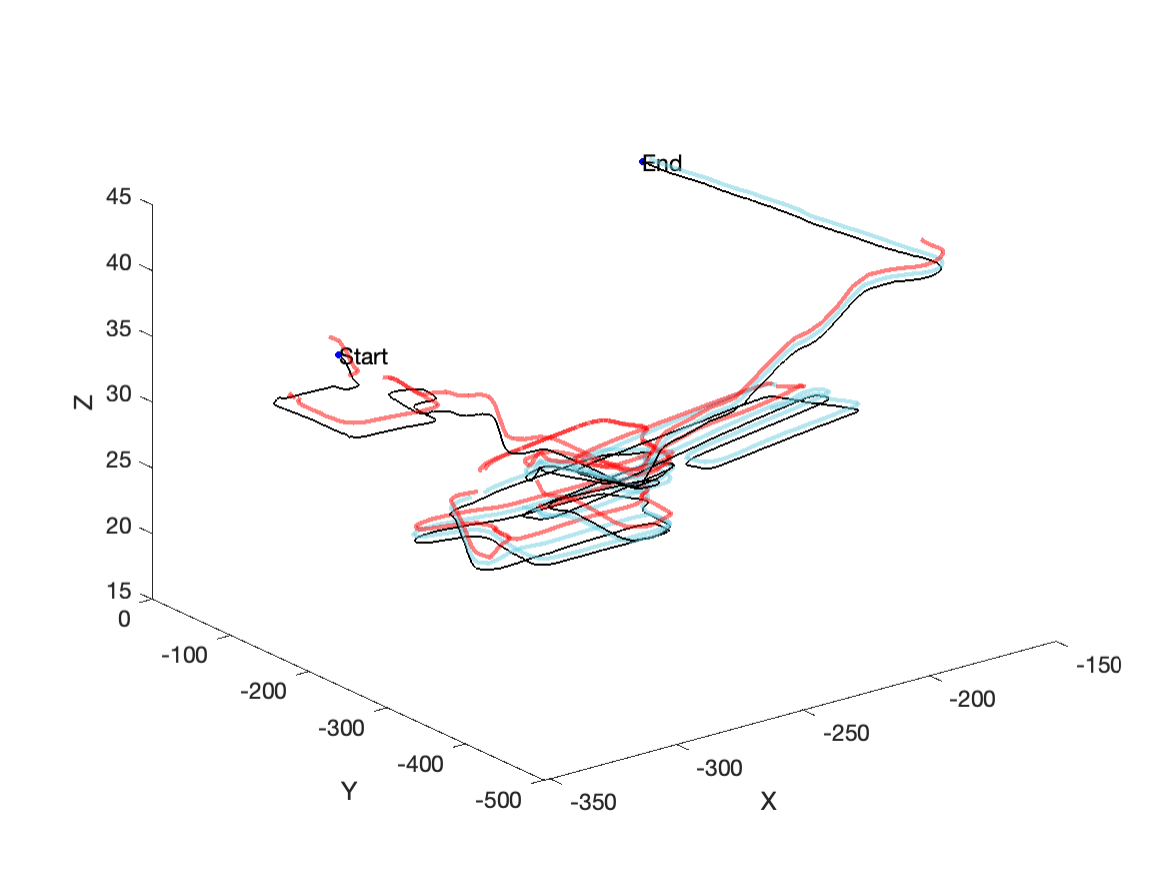}} 
    \subfigure[Lixiangguoji]{\includegraphics[width=0.24\textwidth]{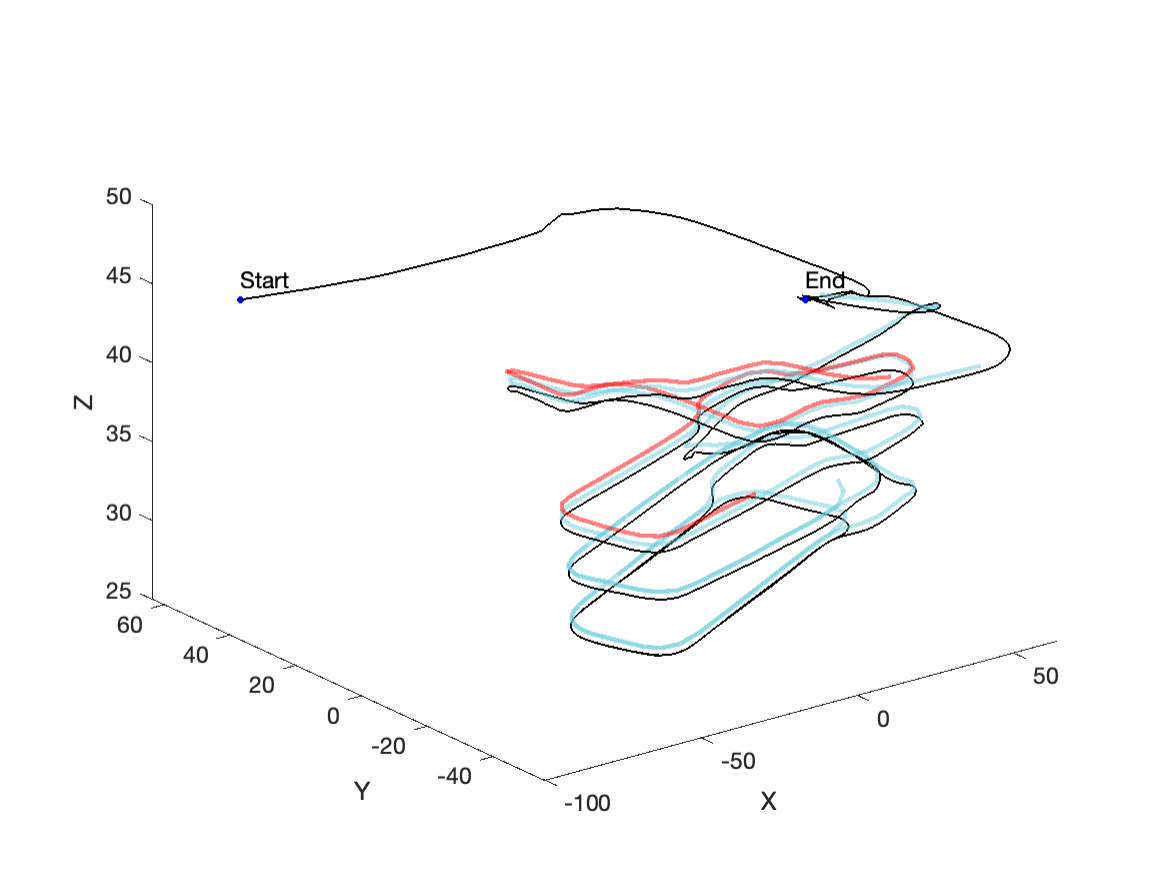}} 
    \subfigure[Yinwang]{\includegraphics[width=0.24\textwidth]{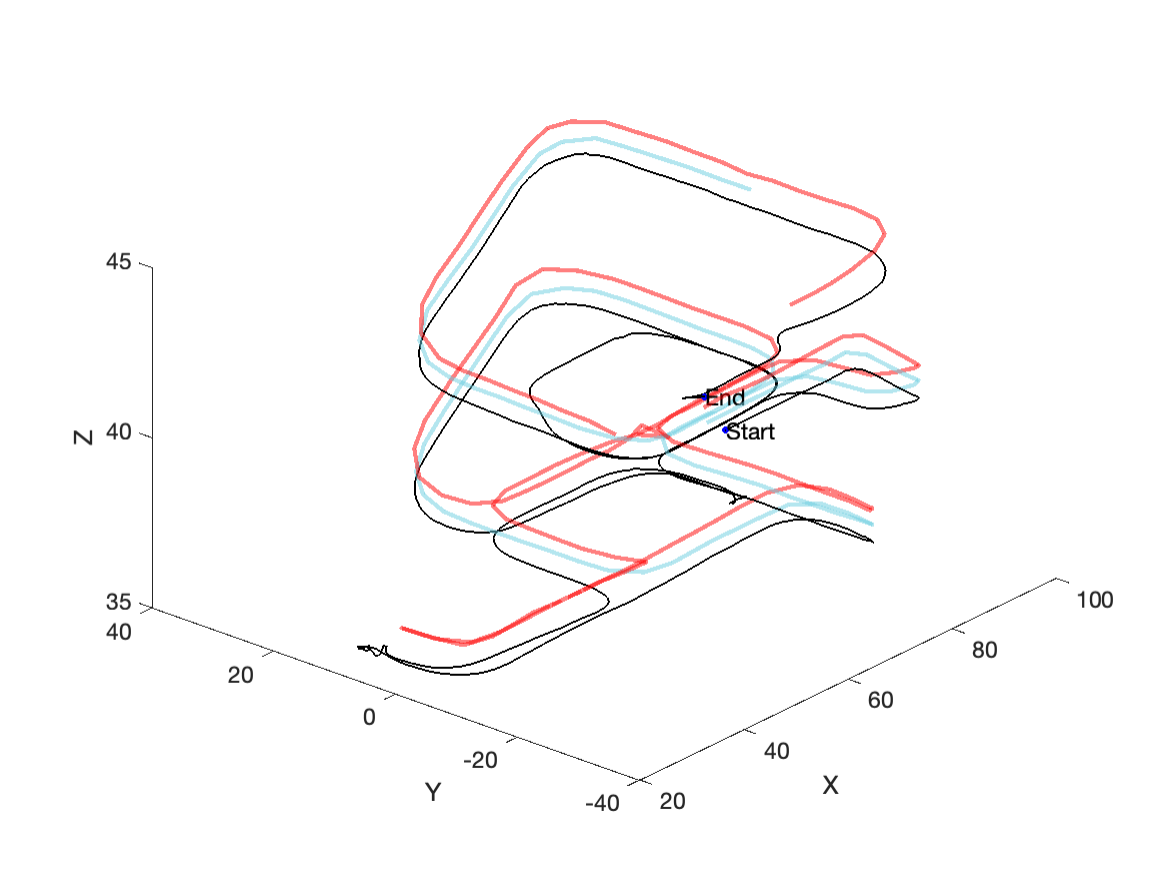}}
    \subfigure[Yinzuo]{\includegraphics[width=0.24\textwidth]{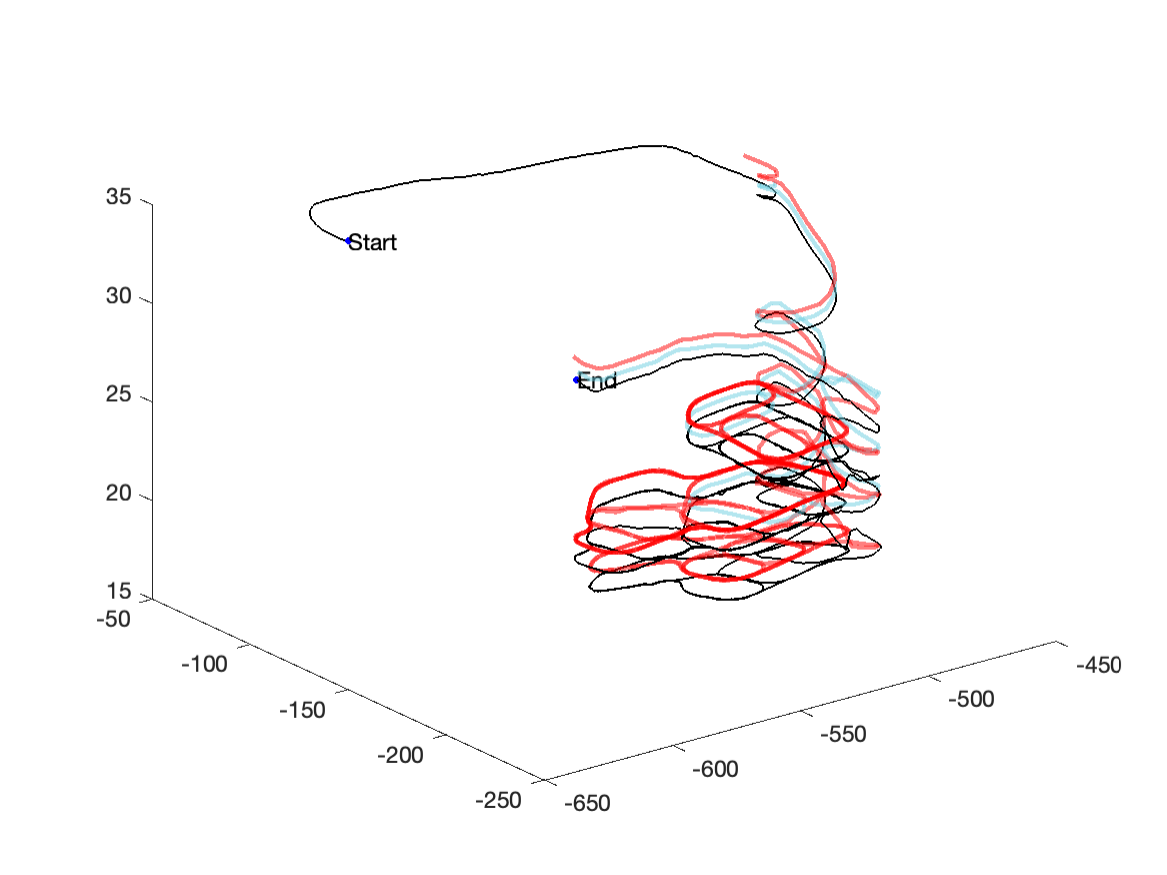}}
    \caption{\kevin{A visualization of trajectories of the experimental dataset. Each sub-figure has three trajectories of different colors, in which blue, red, and black represent the trajectories of test data 1, test data 2, and the database respectively. Trajectories of the same underground parking-lot were collected at different times over different days.}}
    \label{dataset}
\end{figure*}

\kevin{Once the NDT point cloud is obtained, our approach analyzes each NDT cell by categorization of explicit geometric shape types, coupled with calculating the entropy of points within each NDT cell.}

\kevin{Our approach utilizes explicit categories to interpret the shape of each cell. For a given NDT cell, we calculate the sorted eigenvalues $e_1 > e_2 > e_3$ from the covariance matrix, together with their corresponding eigenvectors $v_1, v_2, v_3$. If $e_1 \gg e_2 \approx e_3$,  the shape of the cell will be classified as a straight line. Conversely, if $e_1 \approx e_2 \gg e_3$,  the cell will be considered a plane. However, this shape indicator requires two thresholds: 1) the ratio between $e_1$ and $e_2$, and 2) the ratio between $e_2$ and $e_3$. Instead of employing multiple thresholds, we propose an integrated one-dimensional NDT shape classification index:}
\begin{equation}
\label{g}
g=\frac{e_1\cdot e_3}{\left(e_2\right)^2}
\end{equation}
\kevin{Therefore, by employing a straightforward thresholding strategy on $g$, we can effectively classify the geometric shapes. As illustrated in Fig. \ref{g_figure}, $g$ enables the identification of four distinct shape categories: plane, ellipsoid, sphere, and line. To describe these explicit shape categories, we map the value of $g$ to different shapes using the following formula:}
\begin{equation}
    S = \left\{S \in \mathbb{Z}\ |\ S=\lceil \frac{g}{s} \rceil, 0\le g \le g_{max} \right\}
\end{equation}
\kevin{Here, $g_{\text{max}}$, $s$, and $S$ denote the maximum $g$ value selected, the segmented value we choose, and the geometric value, respectively.}
\kevin{Taking inspiration from NDD \cite{NDD}, we consider each NDT cell as a normal distribution characterized by a mean value $\mu$ and a covariance matrix $\Sigma$. Consequently, we utilize entropy, a statistical measure associated with the normal distribution, to describe the NDT cell:}
\begin{equation}
\label{entropy}
E=\frac{N}{2}(log2\pi+1)+\frac{1}{2} log\left| \Sigma \right|
\end{equation}
where $\left| \cdot \right|$ represents matrix determinant. $N$ represents the dimension of the vector space, and in this particular case, the value of N is 3 because the points involved are in three-dimensional space.

\kevin{Then, the Polar-Range-Height ROI partition $r$, $\theta$, $w$, geometric value $S$, and entropy $E$ of each NDT cell can be obtained. Due to the variance of spatial partition between the voxel grid of NDT and ROI bins, one ROI bin likely contains multiple NDT cells. The majority value of geometric values is used for $S$, and sum-pooling is applied for the entropy $E$.}
\llz{The proposed descriptor NDT-MC integrates both geometric value and entropy features.}

\subsection{Encoding}

\llz{To adapt to both indoor and outdoor scenarios, NDT-MC is devised by combining geometric values $S_{r,\theta,w}$ and entropy $E_{r,\theta,w}$. Specifically, NDT-MC can be obtained by calculating $\mathcal{G}$ and $\mathcal{E}$ for each Region of Interest (ROI):}
\begin{equation}
    \mathcal{G}_{r,\theta}=\sum_{w=0}^{N_z-1}{\left(w+1\right)\cdot}S_{r,\theta,w}
    \label{eq:11}
\end{equation}
\begin{equation}
    \label{mathcal_E_r}
    \mathcal{E}_{r,\theta}=\sum_{w=0}^{N_z-1}{\left(w+1\right)\cdot}E_{r,\theta,w}
\end{equation}
\llz{where $N_z$ represents the number of selected height layers.}


\section{DESCRIPTORS MATCHING}

\subsection{Fast Matching}

\textbf{Construction of geometric key (GK)}: \kevin{Instead of using complete descriptors, we propose a histogram-based key called \emph{geometric key}, to reduce the computation in similarity matching of descriptors. 
As shown in Fig. \ref{GK}, we discrete the continuous $g$ values of NDT representations into columns with an increment of 1. This process results in a $1 \times N_s$ vector, where $N_s$ represents the maximum value of the selected geometric value $S$.}

\textbf{Fast Retrieval based on kd-tree}: 
\kevin{By using the geometry key, the time complexity of kd-tree construction can be reduced from $O(nN_r N_{\theta} \log(n))$ to $O(nN_s \log(n))$, given the size of geometry $N_s$ is much smaller than both $N_r$ and $N_{\theta}$.}
\kevin{Once the kd-tree is built, we undergo the k-nearest neighbor search for each query. }


\textbf{Construction of Sector Key}: If a candidate keyframe satisfies the adaptive distance threshold, \kevin{we compare its descriptor with that of the current frame using a similarity score. As mentioned earlier, rotation invariance is achieved by performing column shifts. However, iterating all columns will inevitably increase the computational complexity. Following $Scan Context$, we compute the column-wise average of the descriptor to obtain a $1 \times N_{\theta}$ vector called the \emph{sector key}.
Sector keys simplify column shift approximation calculation, and descriptor matching is performed in the vicinity of the estimated column shift.}

\begin{table*}[h]\color{black}
\centering
\caption{The comparison of the F1 score and extended precision results on the NIO dataset.}
\label{f1score_nio}
\normalsize
\resizebox{0.98\linewidth}{!}{
\begin{tabular}{c|c|c|c|c|c|c|c|c}
\hline
Method & Rongke1 & Rongke2 & Lixiangguoji1 & Lixiangguoji2 & Yinwang1 & Yinwang2 & Yinzuo1 & Yinzuo2 \\
\hline
NDT-MC & \textbf{0.831}/\textbf{0.521} & \textbf{0.778}/\textbf{0.553} & \textbf{0.894}/\textbf{0.726} & \textbf{0.882}/\textbf{0.509} & \textbf{0.912}/\textbf{0.744} & \textbf{0.927}/\textbf{0.845} & \textbf{0.923}/\textbf{0.628} & \textbf{0.933}/\textbf{0.702} \\
\hline
Scan Context & 0.551/— & 0.360/— & 0.896/— & 0.833/— & 0.747/— & 0.855/— & 0.418/— & 0.913/— \\
\hline
Lidar Iris & 0.641/0.362 & 0.743/0.518 & 0.867/0.566 & 0.878/0.518 & 0.754/0.599 & 0.861/0.525 & 0.743/0.415 & 0.862/0.672 \\
\hline
STD & 0.171/— & 0.061/— & 0.176/— & 0.142/0.511 & 0.139/— & 0.156/— & 0.054/0.503 & 0.136/0.501 \\
\hline
NDT-Hist & 0.116/— & 0.114/— & 0.377/— & 0.330/— & 0.321/— & 0.258/— & 0.296/— & 0.378/— \\
\hline
NDD & 0.611/— & 0.276/— & 0.852/0.635 & 0.866/0.507 & 0.881/0.725 & 0.689/0.399 & 0.843/0.615 & 0.879/0.434 \\
\hline
\end{tabular}
}
\end{table*}

\begin{table*}
\centering
\caption{The comparison of the F1 score and extended precision results on the KITTI dataset.}
\label{f1score}
\tiny
\resizebox{0.98\linewidth}{!}{
\begin{tabular}{c|c|c|c|c|c|c}
\hline
Method & 00 & 02 & 05 & 06 & 07 & 08 \\
\hline
SC & 0.924/0.891 & 0.690/0.516 & 0.859/0.902 & 0.932/0.982 & 0.482/0.630 & 0.608/0.667 \\
\hline
ISC & 0.856/0.737 & 0.675/0.510 & 0.847/0.813 & 0.937/0.921 & 0.506/0.634 & 0.719/0.710 \\
\hline
IRIS & 0.873/0.909 & 0.813/\textbf{0.860} & 0.922/0.925 & 0.936/0.971 & 0.585/0.710 & 0.534/0.665 \\
\hline
M2DP & 0.885/0.911 & 0.616/0.500 & 0.802/0.799 & 0.945/0.920 & 0.515/0.589 & 0.022/0.500 \\
\hline
OverlapTransformer & 0.915/0.842 & 0.801/0.646 & 0.853/0.839 & 0.948/0.915 & 0.438/0.520 & 0.375/— \\
\hline
STD & 0.544/0.549 & 0.394/0.505 & 0.734/0.563 & 0.897/0.578 & \textbf{0.756}/0.639 & 0.603/0.508 \\
\hline
NDD\footnote[5] & 0.943/\textbf{0.963} & 0.851/0.592 & 0.947/0.941 & 0.989/0.976 & 0.659/\textbf{0.713} & \textbf{0.851}/0.661 \\
\hline
NDD\cite{NDD} & 0.943/0.963 & 0.846/0.710 & 0.945/0.934 & 0.996/0.998	 & 0.644/0.733 & 0.896/0.904 \\
\hline
NDT-MC & \textbf{0.954}/0.942 & \textbf{0.871}/0.854 & \textbf{0.952}/\textbf{0.949} & \textbf{0.993}/\textbf{0.993} & 0.615/\textbf{0.713} & 0.736/\textbf{0.752} \\
\hline
\end{tabular}
}
\vspace{-0.18in} %
\end{table*}

\begin{table}[h]\color{black}
\centering
\caption{The runtime performance on KITTI.}
\label{time cost}
\tiny
\resizebox{0.98\linewidth}{!}{
\begin{tabular}{c|c|c}
\hline
Method & Descriptor Extraction(ms) & Query(ms) \\
\hline
STD & 10.214 & 13.399 \\
\hline
SC & 1.199 & 1.798 \\
\hline
IRIS & 5.922 & 1023.213 \\
\hline
NDT-MC & \textbf{0.116} & \textbf{0.161}\\
\hline
\end{tabular}
}
\vspace{-0.18in} %
\end{table}

\subsection{Descriptors matching}

\kevin{Our approach uses correlation coefficient as the similarity metric.}

\kevin{Given two descriptors $D_q=\left\{c_q^0,\ c_q^1,\ldots,c_q^{N_\theta}\right\}$, $D_c=\left\{c_c^0,\ c_c^1,\ldots,c_c^{N_\theta}\right\}$, $c_q^i$ and $c_c^i$ represent the $i^{th}$ column of $D_q$ and $D_c$ respectively. }

\kevin{Given a query descriptor $D_q$ to $\left\{c_q^0,\ c_q^1,\ldots,c_q^{N_\theta},c_q^0,\ c_q^1,\ldots,c_q^{N_ \theta-1}\right\}$,} The calculation method of correlation coefficient between the two descriptors is:
\begin{equation}
g_k\left(D_q^k,D_c\right)=1-\frac{1}{N_\theta+1}\sum_{i=0}^{N_\theta}\left(\frac{(c_q^{i+k}-\overline Q) \cdot (c_c^i - \overline C )}{{\Vert c_q^{i+k} - \overline Q \Vert}\cdot{\Vert c_c^i - \overline C \Vert}}\right)
\label{calculate correlation similirity distance}
\end{equation}
Where $\overline C$ and $\overline Q$ represent the mean of all elements of $D_c$ and $D_q^k$ respectively.

\kevin{Then, column-shifting is employed to find the appropriate yaw angle to match:}
\begin{gather}
g_s\left(D_q,D_c\right)=min\left(g_k\left(D_q^k,D_c\right)\right) \label{g_s} \\
k\in\left\{0,1,\ldots,N_\theta-1\right\}
\label{calculate maximum similirity}
\end{gather}
\kevin{where $D_q^k$ means $D_q$ shifted by $k^{th}$. We determine the similarity between two frame descriptors by calculating the minimum similarity distance obtained through column displacement.}

\section{EXPERIMENT EVALUATION}

\subsection{Dataset for experiments}

\kevin{Our experiments have two distinct scenarios. We collect a dataset for underground parking lots localization (NIO underground parking-lot dataset), while the widely-used KITTI dataset\cite{KITTI} is also used for the evaluation of loop closure detection for on-road driving.}

\textbf{NIO underground parking-lot dataset.} \kevin{This dataset is collected in real-world underground parking lots using the ET7 model of NIO's mass-produced cars, equipped with the Innovusion Falcon LiDAR(that of a FoV of $ 120^{\circ} \times 25^{\circ}$)}. As shown in Fig. \ref{dataset}, the dataset consists of four different parking lots, located in the basements of four commercial malls, i.e. Rongke, Lixiangguoji, Yinwang, and Yinzuo. A NovAtel IMU-ISA-100C system is used to generate ground truth trajectories for loop closures.
A place is defined as a submap built by a trajectory segment of 4m. 
For each underground parking lot, the session with the largest map area (with the largest number of places), is selected as the database. Other sessions are used for testing. Because the pipeline is designed for crowd-sourced mapping, NDT representation is used instead of raw point cloud map. If the distance between the query pose and database pose is within the range of 4m on the x-y plane and 2m on the z-axis to a database pose, it will be considered as a true positive pair. The numbers of loop closures in test data 1 and test data 2 at Rongke are 439 and 415 respectively. Similarly, the numbers of loop closures at Lixiangguoji, Yinwang, and Yinzuo are 310 and 91, 94 and 142, 144 and 454.
\kevin{In Fig. \ref{dataset}, the database and testing trajectories of the four experiments are shown.}

\textbf{KITTI dataset.} \kevin{The sequence data in the KITTI dataset is used for this evaluation, where point cloud data is acquired using the Velodyne HDL-64E lidar sensor. Our experiment follows the same setting with NDD \cite{NDD}.}

\subsection{Experiment on NIO underground parking-lot dataset.} 
In the experiment, the database frames are extracted by cropping the NDT submap across the mapping trajectories. The testing observation is extracted from the local NDT map built on the fly by a Lidar mapper. There are three sessions of data were collected, one for the database and the other for testing.
For each query observation, Eq. \ref{calculate correlation similirity distance} and Eq. \ref{g_s} are used to retrieve the most similar location from the database. Inter-session poses within 4 meters (with z differences not exceeding 2m) will be considered closed loops. In this experiment, the dense point cloud is not applicable. We use the mean values of NDT cells as the input of SOTA methods.

\kevin{The proposed method is evaluated on eight testing datasets collected in four underground parking lots. For comparison, state-of-the-art place recognition methods are implemented, i.e., Scan Context\cite{kim2018scan}, Lidar Iris\cite{wang2020lidar}, NDD\cite{NDD}, STD\cite{yuan2023std} and NDT-Histograms\cite{ndt_histograms}. Since Scan Context-like methods are not designed for underground parking lots, we make the following adaptations: points below 2m w.r.t. lidar coordinate system are used for descriptor representation. For Scan Context and Lidar Iris, the mean values of NDT Submap as the input. For a fair comparison, the proposed method, Scan Context, and Lidar Iris are set with the same parameters, i.e., $N_r$=40, $N_\theta$=60, $Z$=3m, $L_z$=1m and $R$=80m. }Since NDD and STD require the raw point cloud as input, both the observation and the database use the raw point cloud submap stitched by LIO. We use two widely-used evaluation metrics as NDD, namely F1 score, and Extended Precision (EP).

In the NIO parking lot dataset, performance evaluations of methods such as M2DP, OverlapTransformer, and Intensity Scan Context have not been conducted. M2DP exhibits a low Top-1 recall in scenes like Rongke1, Rongke2, and Liyangguoji1 due to its reliance on dense point clouds, while the sparse NDT data in our dataset adversely affects its performance. The OverlapTransformer method, relying on distance images and designed to run on CPU due to the sparsity of NDT point cloud data, is consequently expected to underperform on the NIO dataset. Lastly, the Intensity Scan Context method relies on specific intensity values, and variations in LiDAR sensor intensity values across different hardware platforms may impact its performance.

\subsection{Experiment on KITTI.} \kevin{In this experiment, we define a true positive detection if the distance between the query and matched database frame node is less than 5 meters. Note, that consecutive frames within a certain range will not be considered as positive pair.}

\kevin{During the testing phase on the KITTI dataset,  we evaluate the proposed NDT-MC as loop-closure detection component for a lidar SLAM system. As a comparison,  other existing global descriptors are also evaluated, including Scan Context\cite{kim2018scan}, Intensity Scan Context\cite{wang2020intensity}, Lidar Iris\cite{wang2020lidar}, M2DP\cite{M2DP}, NDD\cite{NDD} and STD\cite{yuan2023std}. Default parameters are used in SC$\footnote{https://github.com/irapkaist/scancontext}$, ISC$\footnote{https://github.com/wh200720041/iscloam}$, IRIS$\footnote{https://github.com/JoestarK/lidar-Iris}$, M2DP$\footnote{https://github.com/LiHeUA/M2DP}$, NDD$\footnote{https://github.com/zhouruihao1001/NDD}$, and STD$\footnote{https://github.com/hku-mars/STD}$. }The same evaluation metrics, i.e. F1 score and Extended Precision (EP), are used in this experiment.

\kevin{In this paper, the parameters of NDT-MC are set as follows: $N_r=20$, $N_{\theta}=60$, $N_w=6$, $g_{max}=2.4$, and the maximum point cloud range is 80m. The proposed descriptor is a $(2\times20)\times60$ matrix. }

\subsection{Evaluation as an Integrated full-SLAM system.} \kevin{Different from most of the state-of-the-art descriptors, our approach can deploy in real-time using a normal CPU. We integrate NDT-MC with LIO-SAM\cite{LIOSAM} and set $K=10$. }
 
Afterward, we compare our method with open-source algorithms SC-LIO-SAM in terms of real-time loop closure performance. we compare the real-time loop closure performance between both our integrated systems, i.e., NDT-MC with LIO-SAM, and SC-LIO-SAM.
For a fair comparison, parameters, i.e. a maximum distance of 80 meters, a similarity distance threshold of 0.6, and a submap update frequency of 1Hz are used in both our approach and the baseline. 
A video demo of the integrated full SLAM system can be accessed through the hyperlink provided below \url{https://youtu.be/xCtWRlEKCfk?si=J-_TAYcQmW1g_Juv}.

\subsection{Run-time performance.} 

We also conduct a comparison of runtime performance for descriptor extraction and retrieval on the KITTI. Our method is benchmarked with Scan Context, Lidar Iris, and STD. All comparisons were performed on a desktop equipped with an i9-10900K CPU @ 3.70GHz and 32GB of memory. As shown in Table\ref{time cost}, our approach shows a supreme performance in terms of both the descriptor extraction and query time.

\subsection{Results analysis}

			

\begin{figure}
    \centering  
    \includegraphics[width=\linewidth]{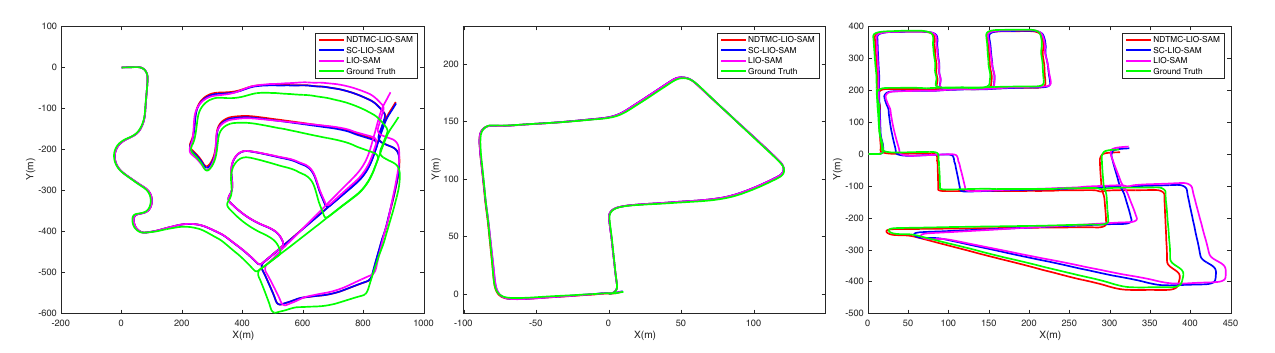} 
    \caption{\kevin{The comparison of trajectories on KITTI sequence 02, 07, and 08. The trajectories of NDTMC-LIO-SAM, SC-LIO-SAM, LIO-SAM, and ground truth are represented in red, blue, rose red, and green, respectively.}}
    \label{trajectorycompare}
    \vspace{-0.18in} %
\end{figure}

TABLE \ref{f1score_nio} presents the F1 Score and EP of the proposed method and the compared methods and NDT point clouds of 2m resolution are investigated. The use of "—" in the table signifies the inability to compute the Extended Precision at various thresholds. This is because these methods did not achieve 100 \% precision at different thresholds, hence preventing the derivation of the $R_{P100}$ (recall at a precision of 100 \% ) value. The Extended Precision (EP) is calculated as $0.5\times(R_{P100}+P_{R0})$, $P_{R0}$ is the precision at minimum recall.
On the eight testing trajectories of four testing scenes, our approach achieves the highest F1 Score and EP. Specifically, our approach with an NDT point cloud resolution achieves the highest performance across all eight test datasets, demonstrating significant advancements compared to other baseline methods.

\llz{NDT-MC presents superior performance in underground parking scenarios due to its effective utilization of height information, its ability to capture complex geometric structures, and its capacity to handle dynamic environments. Encoding and weighting height information in a limited vertical range enhances loop closure accuracy. NDT-MC categorizes NDT cells to capture geometric intricacies, and it emphasizes permanent structure shapes over non-static elements to ensure robust performance in dynamic environments. This approach provides a unique and stable feature identification for prolonged and accurate localization, surpassing methods that rely on less informative high z-values, especially in indoor and dynamic scenes.}

\kevin{TABLE \ref{f1score} presents the comparison results of F1 Score and EP on the KITTI dataset. In the six test scenarios, our method NDT-MC achieved the highest F1 Score in sequences 00, 02, 05, and 06, and the highest EP in sequences 05, 06, 07, and 08. Additionally, our method ranked second in EP in sequences 00 and 02, and second in F1 Score in sequences 07 and 08. Specifically, our method has shown significant advances in the experiment on the KITTI dataset. Our method achieves superior performance on sequences 00, 02, 05, and 06 over the six sequences.}
\llz{In this table, NDD without superscript represents the test results from the NDD paper\cite{NDD}, while NDD with superscript indicates the results obtained by the provided code.}

\llz{We provide standalone comparisons of global descriptors and matching performance on both the NIO dataset and KITTI datasets, as detailed in Sections V-B and V-C. Note that in these comparisons, no prior poses were utilized. Our technique demonstrated superior results on the KITTI dataset, evident from the relatively high F1 Scores and Extended Precision. On the NIO dataset, our method's performance was particularly noteworthy. Additionally, in Section V-D, we provided the assessment results of our proposed technique within an integrated SLAM framework. For this experiment, our proposed global descriptor and matching were combined with the selection of a prior pose for loop-closure detection. We named this system NDTMC-LIO-SAM. Its performance was benchmarked against a leading open-source Lidar SLAM system, SC-LIO-SAM, by comparing overall trajectory accuracy. This is a system-level comparison that takes into consideration integration and run-time factors.}
From Fig. \ref{trajectorycompare}, the proposed NDT-MC-LIO-SAM and the SC-LIO-SAM achieve similar performance on sequences 02 and 07. Especially on sequence 08, the trajectory of our method significantly outperforms SC-LIO-SAM.

\section{CONCLUSION}

\llz{This paper proposed a real-time loop-closure detection approach based on a geometry-only global descriptor, which shows good generalizability in a variety of real-world scenes. In terms of indoor scenes, we used a mass-produced front-view LiDAR for implementation and evaluation. To overcome the limitations of FOV and visibility, we proposed to use an instantly-built SLAM map during the localization process. By this means, spatial-temporal mapping can largely eliminate occlusions and improve co-visibility between databases and query observations. Additionally, our approach leverages a lightweight NDT point cloud representation and encodes explicit geometric shape information by applying shape classification to NDT cells.}

\llz{For localization in both underground parking and on-road driving scenarios, NDT-MC combines the height of cells with their shape and entropy.} Our method is thoroughly tested on eight datasets collected from four indoor underground parking lots and the most widely-used KITTI dataset. Experimental results demonstrate significant advantages of our method in terms of numerous metrics of accuracy and efficiency. We also integrate the proposed global descriptor as a real-time loop closure detection component with a lidar mapping system. Codes are available publicly to the public.








\end{document}